\documentclass[conference]{IEEEtran}
\IEEEoverridecommandlockouts
\usepackage{cite}
\usepackage{amsmath,amssymb,amsfonts}
\usepackage{algorithmic}
\usepackage{graphicx}
\usepackage{textcomp}
\usepackage{xcolor}
\usepackage{booktabs}
\usepackage{tabularx}
\usepackage{array}
\usepackage{xurl}
\usepackage[hidelinks]{hyperref}

\def\BibTeX{{\rm B\kern-.05em{\sc i\kern-.025em b}\kern-.08em
    T\kern-.1667em\lower.7ex\hbox{E}\kern-.125emX}}

\begin{document}

\title{Seeing Less Is Not Seeing Safely: Privacy Leakage from Task-Scoped Robot Perception Exports}

\author{
\IEEEauthorblockN{Yuqiao Xu and Erman Ayday}
\IEEEauthorblockA{
\textit{Department of Computer and Data Sciences} \\
\textit{Case Western Reserve University} \\
Cleveland, OH, USA \\
Yuqiao Xu: your-email@case.edu \\
Erman Ayday: erman.ayday@case.edu
}
}

\author{\IEEEauthorblockN{1\textsuperscript{st} Yuqiao Xu}
\IEEEauthorblockA{\textit{Department of Computer and Data Sciences} \\
\textit{Case Western Reserve University}\\
Cleveland, OH, USA\\
ORCID: 0009-0009-3552-2136}
\and
\IEEEauthorblockN{2\textsuperscript{nd} Erman Ayday}
\IEEEauthorblockA{\textit{Department of Computer and Data Sciences} \\
\textit{Case Western Reserve University}\\
Cleveland, OH, USA\\
ORCID: 0000-0003-3383-1081}
}

\maketitle
\begin{abstract}
Domestic robots rely on rich perception to operate in private homes, but
privacy risk persists even when raw sensor data remain local. Structured
representations exported to downstream planners, cloud services, logs, or
learning pipelines can still reveal household information through semantics,
geometry, spatial structure, and task targets. We introduce
\emph{Task-Functional Perception Distillation} (TFPD), a task-scoped
representation-export framework that keeps rich perception local and profiles
downstream exports according to task utility, direct exposure, and multiple
residual inference risks. Using 120 AI2-THOR scenes with scene-disjoint train/validation/test splits, frozen attacker selection, and representation-aware held-out attacks, we evaluate navigation, collision checking, and object-goal execution. Three
navigation exports achieve identical success ($1.000$) and mean path ratio
($0.898$), yet representation-level linkability ranges from $0.532$ to
$0.970$. Replacing an explicit target label with a target region reduces
target-category macro-F1 from $1.000$ to $0.077$ while preserving success at
$0.995$, while geometric coarsening reduces object-category macro-F1 from
$0.704$ to $0.556$ at a measurable collision-utility cost. A ProcTHOR
replication preserves the navigation task-equivalence/privacy-inequivalence
finding while changing the relative ordering of normalized and topological
exports. These results show that neither field removal nor stronger abstraction
induces a universal privacy ordering and motivate task-specific, multi-risk
evaluation of the complete public representation.
\end{abstract}


\maketitle

\section{Introduction}
\label{sec:introduction}

Domestic robots increasingly operate in private living spaces, where rich perception is necessary for navigation, manipulation, safety, and interaction but can also expose sensitive household information ~\cite{huang2025ultralowres,choi2025pcvs,martinson2024privacyaware}. Privacy risk, however, does not end when raw camera images remain local: derived representations can retain sensitive visual, semantic, and geometric information~\cite{mahler2016privacy}. Modern robot architectures exchange structured messages, maps, object representations, and scene graphs among components for planning, collision checking, logging, cloud computation, and learning~\cite{macenski2022ros2,rosinol2020scenegraphs,kehoe2015cloud}. These representations can encode object and room semantics, geometry, spatial layout, and task goals~\cite{ruizsarmiento2017robothome,rosinol2020scenegraphs} and can therefore disclose household context even without exposing the original RGB or RGB-D stream~\cite{mahler2016privacy,murphy2026vacuum}.

This paper focuses on \emph{environmental-perception privacy}: privacy risks carried by representations derived from a robot's sensing and interpretation of the physical home. Prior privacy-preserving robot systems have primarily reduced exposure at or before visual perception ~\cite{huang2025ultralowres,choi2025pcvs,martinson2024privacyaware}. Such defenses are important, but processed semantic, geometric, and spatial representations can themselves retain information about the underlying environment~\cite{speciale2019privacy,chelani2021lineclouds,pan2019privacyleakage}.

This raises a distinct privacy question:
\begin{quote}
\emph{After a robot has interpreted a private environment, what representation
should be allowed to leave the protected local perception boundary?}
\end{quote}

We introduce \emph{Task-Functional Perception Distillation} (TFPD), a representation-export framework that keeps rich perception local while releasing task-scoped representations to downstream consumers. Navigation can use free-space structure without object identities, collision checking can use obstacle geometry without semantic labels, and object-goal execution can use a reference to a target resolved locally rather than its semantic label.

TFPD distinguishes \emph{direct exposure} from \emph{residual inference}. Removing labels, identifiers, or other explicit fields reduces direct disclosure but does not imply that the remaining representation is private. Geometry can reveal object category, spatial organization can reveal room or household characteristics, topology can provide a scene-specific signature, and target location can remain correlated with target semantics. TFPD therefore profiles candidate representations according to three separate properties: the task utility they preserve, the information they expose directly, and the residual privacy risks that remain inferable from the complete public representation.

We study three representative foundational downstream robot functions spanning complementary privacy channels. \emph{Navigation} requires traversability and spatial connectivity and therefore exposes spatial structure. \emph{Collision checking} requires obstacle geometry and clearance information, creating potential geometric leakage even without semantic labels. \emph{Object-goal execution} requires a target reference and can therefore expose target semantics and task intent. Together, these tasks provide spatial, geometric, and semantic cases for studying downstream environmental-perception privacy.

\begin{figure*}[!t]
    \centering

    \begin{minipage}{0.65\textwidth}
        \centering
        \includegraphics[width=\linewidth]{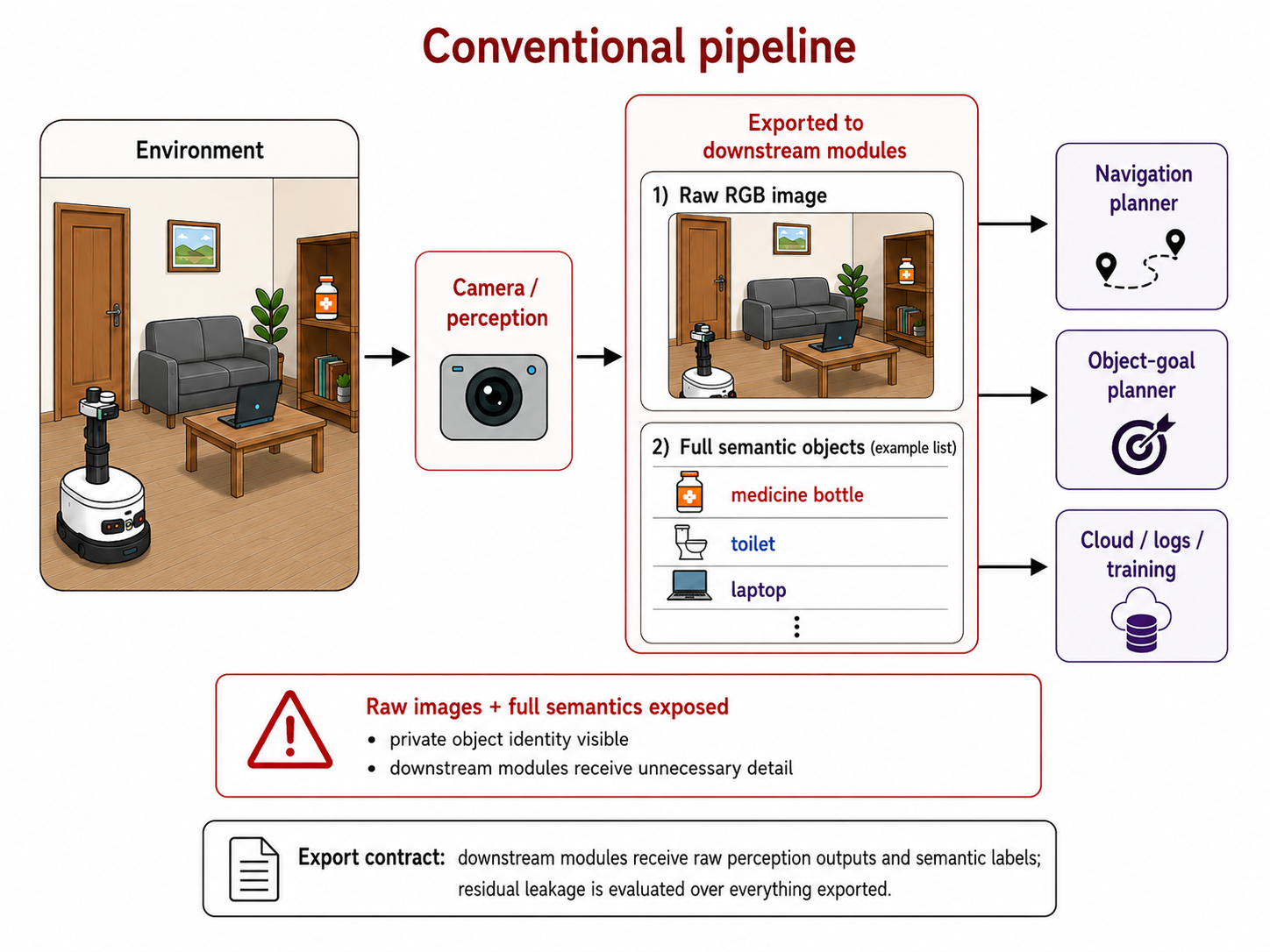}
        \vspace{1mm}
        \textbf{(a) Conventional pipeline}
    \end{minipage}

    \vspace{3mm}

    \begin{minipage}{0.65\textwidth}
        \centering
        \includegraphics[width=\linewidth]{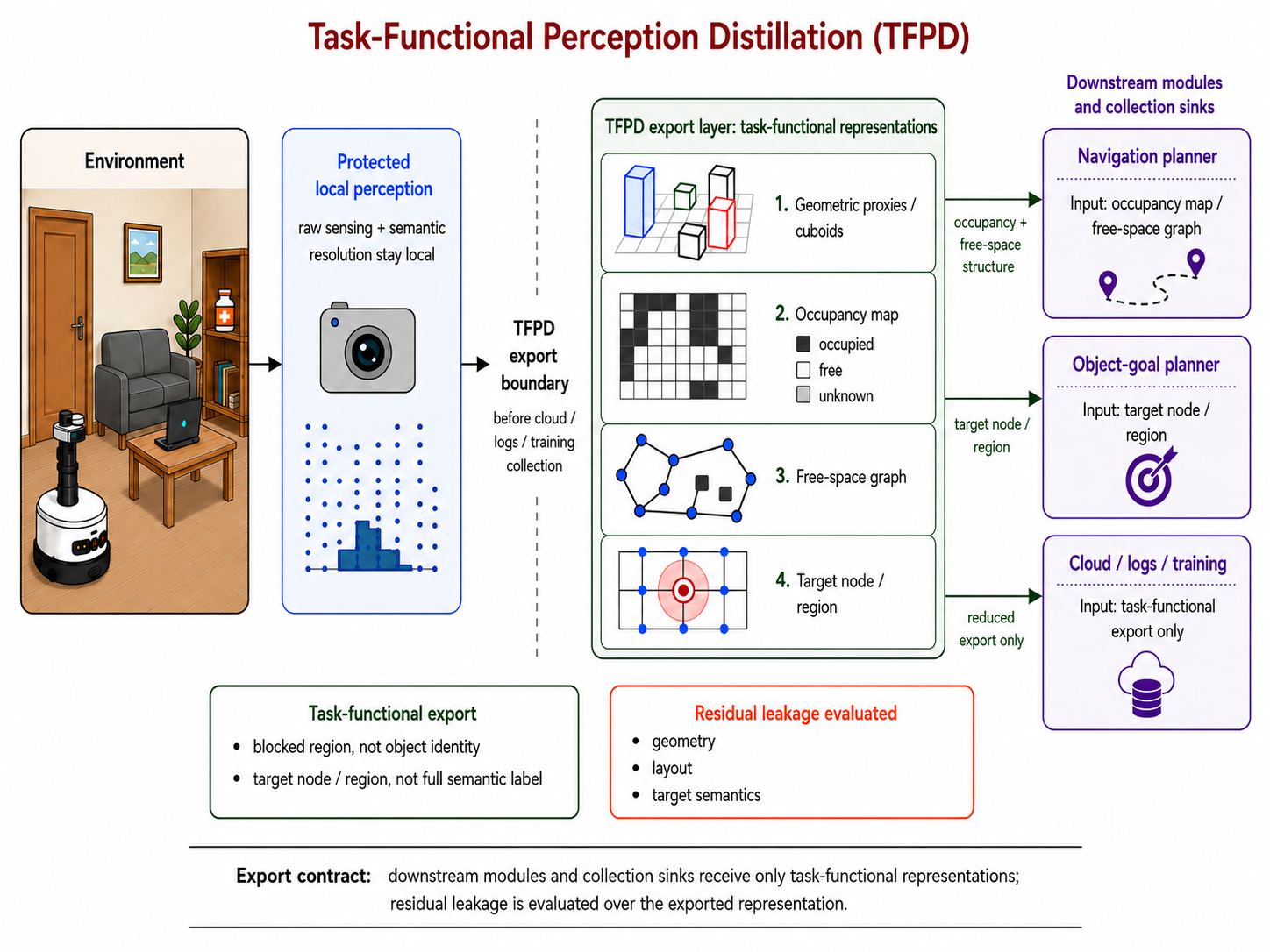}
        \vspace{1mm}
        \textbf{(b) TFPD pipeline}
    \end{minipage}

    \caption{Comparison between a conventional robot perception pipeline and TFPD. A conventional pipeline may expose rich perceptual outputs to downstream planners, logs, cloud services, or learning systems. TFPD places a representation-export boundary after local perception, keeps rich interpretation local, and releases task-scoped representations for downstream use.}
    \label{fig:original-vs-tfpd}
\end{figure*}

As illustrated in Fig.~\ref{fig:original-vs-tfpd}, TFPD inserts a representation-export boundary between rich local perception and downstream consumers. It does not prescribe a universal privacy representation. Instead, each task $\tau$ is associated with a candidate representation family $\mathcal{R}_{\tau}$ whose members may provide equivalent task functionality while exposing different privacy-relevant structure.


Our primary evaluation uses 120 AI2-THOR scenes~\cite{kolve2017ai2thor} with
scene-disjoint training, validation, and held-out test partitions. Attack
models and representation-specific preprocessing are fitted on training data,
selected using validation data, and frozen before held-out evaluation. We
additionally audit 33 representation--attack combinations: only 7 existing
attacks already used the relevant public representation sufficiently, while
the remaining 26 are re-evaluated using broader representation-aware inputs
derived from the same public export. To examine whether the central navigation
finding depends on the original scene distribution, we further repeat the
navigation evaluation on a fixed set of 120 ProcTHOR \cite{deitke2022procthor} houses using the same
representation constructions and evaluation protocol. Because ProcTHOR uses
the AI2-THOR execution engine, we treat this experiment as a
cross-distribution replication rather than validation in an independent
simulator.

The primary AI2-THOR results show that representation abstraction does not
induce a simple monotonic privacy ordering. For navigation, all three
representations are task-equivalent under the evaluated criteria, achieving
success $1.000$, path feasibility $1.000$, and identical mean path ratio
$0.898$, yet representation-level linkability Top-1 ranges from $0.532$ to
$0.970$. Under this primary evaluation, the more abstract topological
representation has higher attack point estimates than coordinate normalization
across all three navigation privacy channels, despite removing more explicit
metric information. For object-goal execution, replacing an explicit target
label with a target region reduces target-category macro-F1 from $1.000$ to
$0.077$ while preserving success at $0.995$. Collision checking exhibits a
different outcome: geometric coarsening reduces object-category macro-F1 from
$0.704$ to $0.556$, while collision F1 decreases from $1.000$ to $0.950$ and
other privacy channels remain strongly inferable.

The ProcTHOR replication reinforces the broader navigation finding while also
showing that the precise privacy ranking is distribution-dependent. Across
1,000 held-out ProcTHOR trials, all three navigation representations again
achieve success and path feasibility of $1.000$ with identical mean path ratio
$0.879$, while linkability ranges from $0.497$ to $1.000$. Metric free space
remains the most linkable representation, but the relative ordering of
normalized and topological free space reverses compared with the primary
AI2-THOR evaluation. Thus, task-equivalent representations remain
privacy-inequivalent across both scene distributions, but stronger abstraction
does not define a universal privacy ordering.

This paper makes the following contributions:
\begin{itemize}

\item \textbf{Post-perception representation privacy surface.}
We identify task-facing environmental representations as a privacy surface
distinct from raw sensing. Even when raw sensor data remain local, exported
representations can disclose household semantics, geometry, spatial structure,
scene-specific information, and task intent.

\item \textbf{Task-Functional Perception Distillation.}
We introduce TFPD, a task-scoped representation-export framework that keeps
rich perception within a protected local boundary and releases downstream
representations constructed around task requirements.

\item \textbf{Representation-aware, multi-risk privacy profiling.}
We distinguish direct disclosure from residual inference and jointly evaluate
candidate exports by task utility and multiple privacy risks. An audit of 33
representation--attack combinations further identifies cases in which
restricted attacks underestimate leakage from information already present in
the public representation.

\item \textbf{Empirical privacy inequivalence across task-functional exports.}
We show that task-equivalent representations can exhibit substantially
different residual privacy risk and that stronger abstraction does not imply a
universal privacy ordering. In the primary AI2-THOR evaluation, navigation
representations with identical measured utility have linkability ranging from
$0.532$ to $0.970$, while collision checking and object-goal execution exhibit
channel-specific privacy improvements, privacy--utility tradeoffs, and
conflicting effects across privacy risks. A cross-distribution replication on
120 ProcTHOR houses preserves the navigation
task-equivalence/privacy-inequivalence finding while changing the relative
ordering of normalized and topological exports.

\end{itemize}

Together, these findings motivate task-specific, multi-risk evaluation of robot
representations rather than treating field removal or increasing abstraction
as evidence of privacy.

\section{Background and Related Work}
\label{Background}

Prior work on privacy in robotics has largely focused on limiting exposure of
raw or visual sensor information. Early work studied privacy--utility tradeoffs
for remotely teleoperated robots by filtering the visual information available
to human operators~\cite{butler2015hri}, while DARKLY explored privacy controls
for camera-based perceptual applications, including robotic systems
~\cite{jana2013darkly}. More recent robot-specific approaches reduce visual
fidelity for privacy-preserving navigation ~\cite{huang2025ultralowres,huang2026userconfigurable}, conceal sensitive
objects or regions in robot video~\cite{choi2025pcvs}, or process images
locally and discard raw observations after extracting information required for
semantic mapping~\cite{martinson2024privacyaware}. Collectively, these
approaches primarily protect information at or before the visual-perception
stage.

Robot systems, however, routinely transform sensor observations into structured
representations that are subsequently consumed by downstream components.
Semantic and object-oriented maps can encode object identities, geometry, and
spatial relationships~\cite{sunderhauf2017meaningful,
ruizsarmiento2017robothome}, while metric-semantic SLAM systems represent
environments using structured geometric and semantic information
~\cite{liu2025slideslam}. In cloud-assisted robotics, perception, inference,
planning, and learning functions can additionally be offloaded to remote
infrastructure~\cite{tanwani2020rilaas}. Thus, keeping raw RGB or depth
observations local does not necessarily prevent derived environmental
information from crossing the robot's local execution boundary.

A closely related line of work protects representations used for particular
perception or robot functions. Privacy-preserving localization transforms scene
geometry or image features while retaining localization utility
~\cite{speciale2019privacy,geppert2021privacy,dusmanu2021privacyfeatures,
shibuya2020privacyslam,moon2024rayclouds}. Private Multiparty Perception learns
privacy-preserving multi-view representations for autonomous navigation
~\cite{lu2022private}. Antonazzi et al. learn task-specific representations for
cloud-based robot object detection while suppressing sensitive information
~\cite{antonazzi2025privacy}, and RoboShape learns privacy-aware point-cloud
representations that preserve object-level utility while reducing inference of
sensitive spatial attributes~\cite{baser2026roboshape}. These works establish
that changing the representation can reduce disclosure while preserving a
particular robot or perception function. TFPD builds on this general insight
but considers the representation exported to different downstream robot tasks
rather than designing one privacy-preserving representation for a single
function.

Representation transformation, however, does not by itself eliminate residual
leakage. Line-cloud representations can retain sufficient spatial structure to
recover approximate scene geometry~\cite{chelani2021lineclouds}, and later
work continues to identify exploitable structure in geometrically obfuscated
representations~\cite{kim2026dcl}. Higher-level structure can also remain
informative: navigation policies can reveal properties of their training
environments, including floor-plan information
~\cite{pan2019privacyleakage}, while room semantics can be inferred from
floor-plan structure~\cite{paudel2021roomclassification}. These findings show
that removing explicit semantics or selected sensitive attributes does not
eliminate the information carried by the remaining geometry, topology, or
spatial organization. \textbf{Attribute-centric transformation is therefore not
sufficient as a general robot-privacy strategy; privacy must also account for
the information required by a robot task and the residual information encoded
by the representation that remains.}

This observation also connects TFPD to the broader principle of data
minimization. Tran and Fioretto study how machine-learning inference can be
performed using smaller subsets of input features
~\cite{tran2023dataminimization}, while Ganesh et al. show that minimizing
collected attributes does not necessarily yield the expected privacy benefit
and motivate considering multiple privacy risks
~\cite{ganesh2025dataminimization}. TFPD shares the objective of limiting
unnecessary disclosure, but differs in its unit of analysis. Rather than
selecting a smaller subset of independent input attributes, TFPD considers the
structured semantic, geometric, and spatial representations required by
different robot tasks. Because information removed explicitly can remain
inferable from the structure that is retained, minimizing individual fields
alone is insufficient; the complete task-facing representation must be
evaluated for residual privacy leakage.

TFPD therefore focuses on the \emph{post-perception representation-export
boundary}. Rather than proposing one privacy transformation for a single robot
function, TFPD considers alternative task-functional exports for navigation,
collision checking, and object-goal execution. Navigation is evaluated through
metric, normalized, and topological free-space representations; collision
checking through anonymous exact and coarsened geometry; and object-goal
execution through alternative semantic, spatial, and structural target
references. Each candidate is profiled according to task utility, direct
exposure, and multiple residual inference risks using information available in
the public export. To our knowledge, prior work has not jointly evaluated,
across multiple downstream robot functions, whether alternative task-functional
representations with the same evaluated task outcome expose comparable privacy
risk, whether stronger abstraction necessarily reduces residual leakage, and
whether reducing one privacy channel also reduces others.

\section{Threat Model and Problem Formulation}
\label{Threat Model}

\textbf{Scope and trust boundary.}
We consider a domestic robot operating in a private living environment. The
robot may use rich sensing modalities such as RGB, RGB-D, LiDAR, or other
onboard sensors, while its local perception stack performs semantic
recognition, SLAM, mapping, localization, and target resolution. TFPD does not
restrict the information available to these local functions; instead, it
controls which derived representation is allowed to leave the protected local
perception boundary.

Our scope is \emph{environmental-perception privacy}: privacy risks carried by
representations derived from sensing and interpreting the physical home. We
study navigation, collision checking, and object-goal execution. Privacy risks
arising from human--robot interaction channels such as raw audio, speaker
identity, speech transcripts, and conversational history are outside the scope
of this work.

We assume that the local sensing and perception stack is trusted. Downstream
consumers may include planners, cloud services, logging and debugging systems,
third-party components, and learning pipelines that receive, process, store, or
reuse exported representations. TFPD must therefore operate before protected
perception information is released to these systems; it cannot retroactively
remove exposure that has already occurred.

\textbf{Adversary.}
The adversary observes the complete representation exported across the TFPD
boundary. Depending on the task, this may include free-space structure,
anonymous or coarsened obstacle geometry, a target coordinate or region, an
anonymous target identifier, or a graph-based representation. The adversary
may represent a curious or compromised downstream service or an analyst with
access to stored exports, but does not observe raw sensor streams, hidden
semantic state, or fields retained exclusively inside local perception.

We conservatively assume that the adversary knows the schema of the exported
representation, the downstream task, and the public transformation used to
construct it. This is a threat-model assumption rather than a claim that every
attacker necessarily has such knowledge. A service that legitimately consumes
the export must understand its interface, and a curious or compromised
downstream component may therefore obtain the same representation schema or
transformation specification. TFPD consequently does not rely on keeping the
representation format or transformation secret.

Attack models and representation-specific preprocessing are fitted using
training data, selected using validation data, and frozen before held-out test
evaluation. The attacker receives no additional information from inside the
protected local boundary.

\textbf{Residual privacy risks.}
We evaluate five representative inference risks:
\begin{itemize}
\item \textbf{Room inference:} inferring room or environmental context from the
exported representation;

\item \textbf{Private-object inference:} inferring whether privacy-sensitive
household objects are present;

\item \textbf{Object-category inference:} recovering semantic object categories
from anonymous or coarsened geometry;

\item \textbf{Scene linkability:} matching a perturbed exported representation
to the corresponding scene in a gallery. Because the query is derived from a
perturbation of the same underlying export rather than an independent
cross-session recapture, this measures representation-level linkability rather
than real-world cross-session home re-identification; and

\item \textbf{Target-category inference:} recovering the semantic category of an
object-goal target from its exported representation.
\end{itemize}

These attacks measure \emph{residual inference}: information that is not
explicitly disclosed may nevertheless remain recoverable from geometry,
relative position, spatial layout, topology, task context, or other structure
contained in the public export.

\textbf{Privacy and utility goals.}
TFPD has three goals. First, it reduces \emph{direct exposure} by withholding
raw images, complete semantic maps, object identities, target labels, or other
fields that are unnecessary for a downstream task. Second, it characterizes
\emph{residual inference} from the complete public representation rather than
assuming that removing explicit fields provides privacy. Third, it preserves
\emph{task utility}, which we evaluate using navigation success and path
quality, collision-checking performance, and object-goal success. This allows
us to distinguish a privacy improvement without measured utility loss from a
privacy--utility tradeoff or a residual conflict across privacy channels.

\textbf{Non-goals.}
TFPD does not protect against compromise of the trusted local perception stack
and does not provide a formal guarantee that exported representations reveal no
private information. Measured attack performance characterizes residual
leakage under the evaluated adversaries rather than an upper bound on all
possible inference. We also do not introduce a new navigation planner,
collision-checking algorithm, or robot controller, nor do we perform adaptive
runtime selection among candidate representations.

\section{Task-Functional Perception Distillation}
\label{sec:framework}

Task-Functional Perception Distillation (TFPD) controls what information derived
from rich local perception is released to downstream robot components. It does
not restrict perception inside the trusted local stack. Instead, TFPD constructs
a task-scoped representation before information crosses the
representation-export boundary and profiles that representation according to
task utility, direct exposure, and residual inference risk.

\subsection{Task-Scoped Export Construction}

Let $S$ denote a private home scene and $P(S)$ the rich perception state
maintained inside the protected local boundary; for brevity, we write
$P=P(S)$. Depending on the robot, $P$ may contain sensor-derived observations,
semantic object information, object identities, detailed geometry, free-space
structure, localization state, and resolved target information.

For downstream task $\tau$, TFPD defines a family of candidate exports
\[
\mathcal{R}_{\tau}
=
\{R_{\tau,1},\ldots,R_{\tau,m}\},
\]
where each candidate is constructed from the same local perception state:
\[
R_{\tau,j}=D_{\tau,j}(P).
\]
Candidates for the same task may satisfy the same functional requirement while
exposing different semantic, geometric, or spatial structure.

TFPD characterizes each transformation through three information-handling
roles. Let $\mathcal{F}(P)$ denote the information available in the local
perception state. For candidate $R_{\tau,j}$, let
$\mathcal{K}_{\tau,j}\subseteq\mathcal{F}(P)$ denote information retained and
exposed directly, and let
$\mathcal{X}_{\tau,j}\subseteq\mathcal{F}(P)$ denote information used locally
to construct a transformed task-functional representation. Information that is
neither exposed nor used to construct the export is
\[
\mathcal{O}_{\tau,j}
=
\mathcal{F}(P)
\setminus
\left(
\mathcal{K}_{\tau,j}
\cup
\mathcal{X}_{\tau,j}
\right).
\]

Let $\pi_{\mathcal{A}}(P)$ denote projection of $P$ onto information
$\mathcal{A}$. A candidate export is represented as
\[
R_{\tau,j}
=
\left[
\pi_{\mathcal{K}_{\tau,j}}(P),
\;
G_{\tau,j}
\left(
\pi_{\mathcal{X}_{\tau,j}}(P)
\right)
\right],
\]
where $G_{\tau,j}$ constructs a task-functional replacement from locally
available information. Pure field removal is the special case
$\mathcal{X}_{\tau,j}=\varnothing$.

\begin{figure*}[t]
\centering
\includegraphics[width=\textwidth]{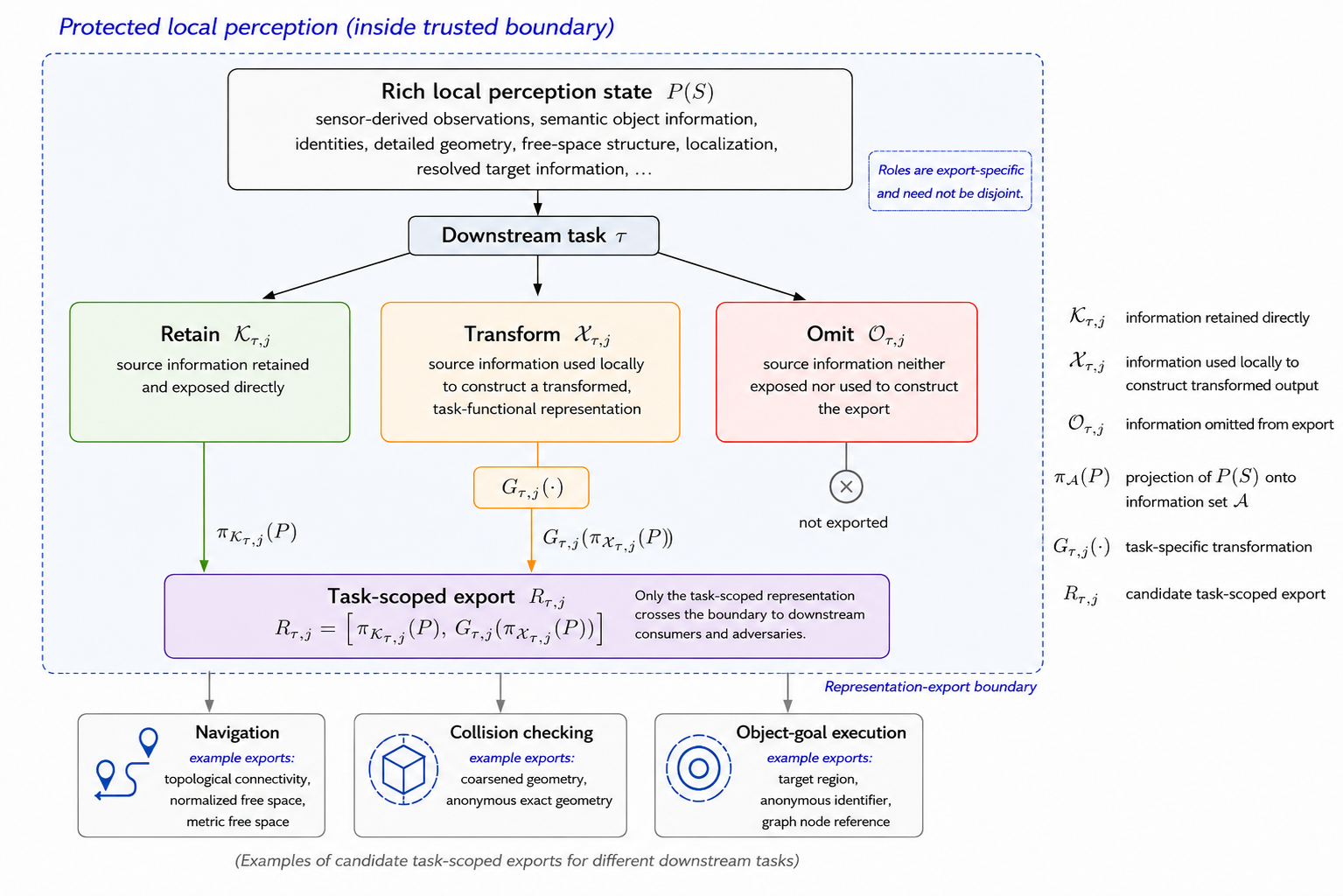}
\caption{TFPD representation construction. For downstream task $\tau$, rich
local perception $P(S)$ is mapped to candidate export $R_{\tau,j}$. Source
information may be retained directly, used locally to construct a transformed
task-functional representation, or omitted. Only the resulting task-scoped
representation crosses the export boundary.}
\label{fig:tfpd-construction}
\end{figure*}

This formulation permits transformation rather than only deletion. Metric
coordinates can be replaced by normalized coordinates, metric free-space
geometry by topological connectivity, detailed obstacle geometry by a
coarsened proxy, and a semantic target resolved locally by a spatial or
structural reference. Importantly, information used locally need not itself
appear in the export: a semantic target category can, for example, be used to
derive a target region while the category remains inside the protected
boundary. These retain, transform, and omit pathways and their relationship to
the representation-export boundary are illustrated in
Fig.~\ref{fig:tfpd-construction}.

\textbf{Public export contract.}
A candidate name such as ``target region,'' ``coarsened geometry,'' or
``topological free space'' identifies the privacy-relevant component varied by
the transformation; it does not imply that this component is the only
information available downstream. Any task-common information required by the
corresponding downstream interface remains part of the complete public export.
Consequently, both utility evaluation and residual privacy analysis operate on
the complete task-facing representation rather than only on the transformed
field. Table~\ref{tab:tfpd-candidates} summarizes these task-specific
functional requirements, candidate representations, and privacy-relevant
transformations. The representation-aware adversary in
Section~\ref{Evaluation} may use all information contained in the public export
but receives no information retained exclusively inside the protected local
boundary.

This distinction is particularly important for object-goal execution. For
example, ``target region'' describes how the already-resolved goal is referenced
downstream; any additional public spatial or planning structure required by the
evaluated task remains part of the task-facing representation. Reducing
target-category inference therefore does not imply that the complete export is
uninformative about other household attributes.

Table~\ref{tab:tfpd-candidates} summarizes the candidate families evaluated in
this work.

\begin{table*}[t]
\centering
\caption{Task-scoped candidate representation families evaluated by TFPD. The
candidate representations differ in the privacy-relevant information exposed
for the corresponding downstream function; task-common public information
remains part of the complete export.}
\label{tab:tfpd-candidates}
\begin{tabular}{
p{0.16\textwidth}
p{0.24\textwidth}
p{0.34\textwidth}
p{0.20\textwidth}}
\toprule
\textbf{Task} &
\textbf{Functional requirement} &
\textbf{Candidate task-facing representations} &
\textbf{Privacy-relevant transformation} \\
\midrule

Navigation &
Traversability, connectivity, and route structure &
Metric, coordinate-normalized, and topological free-space structure &
Remove object semantics; transform or remove absolute metric structure \\

\addlinespace

Collision checking &
Obstacle extent and clearance &
Anonymous exact geometry and coarsened geometry &
Remove semantic identity; optionally reduce geometric precision \\

\addlinespace

Object-goal execution &
Reference to a locally resolved target &
Target label, coordinate, region, anonymous identifier, and graph node &
Replace explicit target semantics with spatial or structural references \\
\bottomrule
\end{tabular}
\end{table*}

\subsection{Task-Specific Instantiations}

\textbf{Navigation.}
Navigation requires traversable-space structure and connectivity. We consider
\[
\mathcal{R}_{\mathrm{nav}}
=
\{
R_{\mathrm{metric}},
R_{\mathrm{normalized}},
R_{\mathrm{topological}}
\}.
\]
The metric representation retains reachable free space in a global metric
frame, including absolute position, scale, and spatial extent. The normalized
representation transforms the same free-space geometry into a local normalized
frame, removing global origin and absolute metric scale while retaining
relative geometry and connectivity. The topological representation replaces
explicit metric geometry with node--edge connectivity, omitting coordinates
and metric edge lengths while retaining reachability. Semantic object labels
and identities are omitted from all three navigation exports.

\textbf{Collision checking.}
Collision checking requires obstacle geometry sufficient for intersection and
clearance reasoning. We consider
\[
\mathcal{R}_{\mathrm{coll}}
=
\{
R_{\mathrm{exact}},
R_{\mathrm{coarsened}}
\}.
\]
Anonymous exact geometry removes semantic labels and identities while retaining
obstacle position and dimensions. Coarsened geometry further transforms this
information into lower-precision obstacle proxies while retaining approximate
extent for collision reasoning. The exact coarsening parameters used in the
evaluation are reported in Section~\ref{Evaluation}.

\textbf{Object-goal execution.}
An object-goal request $q_t$ is first resolved locally:
\[
y_t=H(P,q_t),
\]
where $H$ is the local target-resolution function and $y_t$ is the resolved
semantic target. We consider
\[
\mathcal{R}_{\mathrm{goal}}
=
\{
R_{\mathrm{label}},
R_{\mathrm{coordinate}},
R_{\mathrm{region}},
R_{\mathrm{ID}},
R_{\mathrm{node}}
\}.
\]
The target-label representation exports $y_t$ directly and serves as the
semantic-reference baseline. For the transformed candidates, $y_t$ remains
local and is used to construct a target coordinate, region, anonymous
identifier, or planning-graph node. Thus, the downstream component can receive
a functional reference to the already-resolved target without necessarily
receiving its semantic category.

Across the three tasks, TFPD follows the same principle: retain information
that must remain explicit, transform information when a task-functional
replacement is sufficient, and omit information unnecessary for the selected
interface. TFPD specifies representation construction rather than online
representation selection. Once candidate $j^\star$ is selected for task
$\tau$,
\[
P(S)
\xrightarrow{D_{\tau,j^\star}}
R_{\tau,j^\star}
\longrightarrow
\text{downstream consumer}.
\]
Only $R_{\tau,j^\star}$ crosses the representation-export boundary.

\subsection{Privacy and Utility Profiling}
\label{sec:privacy-profile}

TFPD does not treat field removal or representation transformation as evidence
of privacy. Candidate exports are instead evaluated according to direct
exposure, residual inference, and task utility.

\noindent\textbf{Direct exposure.}
For candidate $R_{\tau,j}$, let $\mathcal{A}_{\tau,j}$ denote the information
permitted by its downstream interface. A conforming export satisfies
\[
\mathcal{F}(R_{\tau,j})
\subseteq
\mathcal{A}_{\tau,j}.
\]
Let $\mathcal{F}_{\mathrm{priv},\tau}$ denote fields treated as directly
privacy-bearing for task $\tau$. Direct exposure is
\[
E_{\tau}(R)
=
\mathcal{F}(R)
\cap
\mathcal{F}_{\mathrm{priv},\tau}.
\]
If
\[
\mathcal{A}_{\tau,j}
\cap
\mathcal{F}_{\mathrm{priv},\tau}
=
\varnothing,
\]
then those fields are not explicitly available through the interface. This is
a construction-level direct non-disclosure property; it does not imply that
the corresponding private information cannot be inferred.

\noindent\textbf{Residual inference.}
For privacy channel $q$, let $Z_q$ denote the private attribute and
$A_{\tau,q}$ an attacker operating on the complete public representation $R$:
\[
\widehat{Z}_q=A_{\tau,q}(R).
\]
Measured residual inference is
\[
L_{\tau,q}(R)
=
\operatorname{Perf}_{q}
\left(
A_{\tau,q}(R), Z_q
\right),
\]
where larger values indicate greater attacker success. We use macro-F1 for
classification attacks and Top-1 accuracy for representation-level
linkability.

Each result is interpreted relative to a non-informative baseline
$B_{\tau,q}$. For classification attacks, the baseline is a
training-set most-frequent-class predictor evaluated using the same macro-F1
metric as the corresponding attacker. The prediction is fixed from training
labels and is not included in attacker selection. For Top-1 linkability with
gallery $\mathcal{G}$, the random-matching baseline is
\[
B_{\tau,q}
=
\frac{1}{|\mathcal{G}|}.
\]

Because the same representation can behave differently across privacy
channels, we retain residual inference as a vector:
\[
\mathbf{L}_{\tau}(R)
=
[
L_{\tau,1}(R),\ldots,L_{\tau,k}(R)
].
\]
Thus,
\[
E_{\tau}(R)=\varnothing
\;\not\Rightarrow\;
\mathbf{L}_{\tau}(R)=\mathbf{0}.
\]
An interface can therefore eliminate direct disclosure of selected fields while
geometry, topology, relative position, spatial layout, or task context still
supports inference about those or other private attributes.

\noindent\textbf{Multi-risk privacy--utility profile.}
Let $U_{\tau}(R)$ denote the measured utility of representation $R$ for task
$\tau$. TFPD characterizes each candidate by
\[
\Phi_{\tau}(R)
=
\left(
U_{\tau}(R),
E_{\tau}(R),
\mathbf{L}_{\tau}(R)
\right).
\]
The components of $\mathbf{L}_{\tau}(R)$ remain separate because the same
transformation can reduce one privacy risk while leaving another unchanged or
increasing it. Collapsing these risks into a scalar would additionally require
application-specific weights that TFPD does not assume.

We further ask whether representations satisfying the same task-utility
criterion also expose comparable privacy. When
\[
U_{\tau}(R_a)=U_{\tau}(R_b),
\]
we write
\[
R_a\equiv_{\tau}R_b.
\]
Task equivalence does not imply privacy equivalence:
\[
R_a\equiv_{\tau}R_b
\;\not\Rightarrow\;
\mathbf{L}_{\tau}(R_a)
=
\mathbf{L}_{\tau}(R_b).
\]
When multiple utility criteria are used, equivalence refers to the complete set
of reported criteria rather than a single metric. In our navigation evaluation,
for example, the three candidate exports have identical success, path
feasibility, and mean path ratio and are therefore task-equivalent under the
evaluated navigation criteria.

\noindent\textbf{Representation-aware evaluation.}
Residual inference is measured from the complete public representation
available to the downstream consumer. Consistent with
Section~\ref{Threat Model}, neither the representation schema nor the public
transformation is treated as secret. Representation-aware attackers may exploit
semantic, geometric, spatial, or structural information contained in the
public export, but receive no information retained exclusively inside the
protected local boundary.

Privacy changes are interpreted separately for each channel. A
\emph{channel-specific privacy improvement} occurs when measured attack success
decreases while task utility is preserved. A \emph{privacy--utility tradeoff}
occurs when reduced attack success accompanies reduced task performance. A
\emph{residual privacy conflict} occurs when one privacy channel improves while
another remains high or changes differently. TFPD does not assign weights
across these heterogeneous risks or automatically select a globally optimal
representation.

\section{Evaluation}
\label{Evaluation}

We evaluate TFPD across three complementary experimental settings with
distinct roles. AI2-THOR serves as the primary evaluation environment and
provides the main evidence for our privacy--utility claims across navigation,
collision checking, and object-goal execution. We additionally repeat the
navigation evaluation on a fixed ProcTHOR house distribution to test whether
the central task-equivalence/privacy-inequivalence finding persists under a
separately generated set of complete, potentially multi-room household
environments. ProcTHOR uses the AI2-THOR execution engine and is therefore
treated as a cross-distribution replication rather than an independent
simulator. Finally, a separately generated synthetic environment is used as a
controlled diagnostic setting to isolate selected leakage mechanisms under
simplified representation changes. Results from these settings are interpreted
within their respective protocols rather than pooled numerically.

The evaluation addresses four questions:
\begin{itemize}

\item \textbf{Q1: Direct disclosure versus residual inference.}
Does eliminating explicit privacy-bearing fields from a downstream
representation also eliminate residual inference about the private environment?

\item \textbf{Q2: Task equivalence versus privacy equivalence.}
Can representations that preserve equivalent measured task utility nevertheless
expose different amounts or types of private information?

\item \textbf{Q3: Information removal versus privacy.}
Does removing more semantic, geometric, or metric information necessarily
produce a more privacy-preserving representation?

\item \textbf{Q4: Representation-aware adversaries.}
How do the privacy conclusions change when attackers make broader use of the
public fields and structure contained in each representation rather than
selected subsets or summaries?

\end{itemize}

The results answer Q1--Q3 negatively and show that representation-aware
evaluation is essential for Q4. The main task-level conclusions persist under
the final representation-aware evaluation, but broader use of the public export
can substantially change individual leakage estimates and even their relative
ordering. For example, metric-navigation private-object inference increases
from $0.405$ under the restricted mapping to $0.928$ under the
validation-selected representation-aware mapping. Overall, removing explicit
private fields does not eliminate residual inference, task-equivalent
representations can remain privacy-inequivalent, and stronger abstraction does
not necessarily yield lower leakage. At the same time, task-specific
transformations can provide channel-specific benefits: coordinate normalization
reduces navigation linkability without measured utility loss, functional target
references sharply reduce target-category inference, and geometric coarsening
reduces object-category inference at a measurable cost in collision-checking
utility. The ProcTHOR replication further shows that the central navigation
privacy-inequivalence result persists across a distinct household-scene
distribution even though the precise ordering of normalized and topological
linkability changes.

\subsection{Primary AI2-THOR Evaluation Protocol}

We use all 120 household scenes in AI2-THOR~\cite{kolve2017ai2thor} with a
fixed scene-disjoint split of 80 training, 20 validation, and 20 held-out test
scenes. The validation and test sets each contain five scenes from each of the
four room families. All representations derived from the same underlying scene
remain in the same split.

Representation-specific preprocessing and attacker models are fitted using
training scenes only. Public feature mappings, candidate attacker families,
and retrieval metrics are selected using validation data and frozen before
held-out test evaluation. The test set is not used for attacker selection,
feature selection, retrieval-metric selection, or class eligibility.

For target-category inference, eligible classes are frozen before test
evaluation using a minimum training count of 10 and presence in the validation
split. Held-out test labels are not inspected when defining the eligible class
set.

We report 95\% confidence intervals from 1,000 scene-cluster bootstrap
replicates. These intervals characterize uncertainty over the evaluated scenes
and observations associated with those scenes and are not interpreted as
formal pairwise significance tests.

\paragraph{Evaluation units and utility.}
Evaluation units depend on the privacy channel. Navigation room and
private-object inference, and collision room and private-object inference, use
the 20 held-out scenes. Collision object-category inference is evaluated over
848 held-out object records. Object-goal room and private-object inference use
400 held-out episodes, while target-category inference uses 328 eligible
episodes after the frozen class-eligibility rule described above.

Navigation utility is evaluated using 50 deterministic start--goal trials per
held-out scene, yielding 1,000 trials. We measure task success, path
feasibility, and mean path ratio relative to the original reachable-space
graph. All three navigation representations achieve success $1.000$, path
feasibility $1.000$, identical mean path ratio $0.898$
[95\% CI: $0.890,0.907$], and zero held-out planning failures. Collision
utility is measured using collision F1. Object-goal utility is measured using
execution success after local target resolution.

\paragraph{Classification attacks and non-informative baselines.}
All classification targets are single-label; private-object inference is a
binary presence task. Candidate attacker families include scaled logistic
regression, random forests, extra trees, and histogram gradient boosting, with
the feature mapping and model family selected using validation macro-F1.

Classification results are interpreted relative to a training-set
most-frequent-class predictor evaluated using the same macro-F1 metric. This
baseline is fixed from training labels and is excluded from attacker selection.
For navigation, the room and private-object baselines are $0.100$ and $0.444$,
respectively. Collision checking uses the same scene-level baselines, while
the object-category baseline is $0.00160$ over 848 held-out objects; 92
classes are frozen as eligible before test evaluation and 90 occur in the
held-out labels. For object-goal execution, the episode-level room and
private-object baselines are again $0.100$ and $0.444$, while the
target-category baseline is $0.00482$ over 328 eligible episodes and 52 frozen
target classes.

\paragraph{Representation-level linkability.}
One representation from each of the 20 held-out scenes forms the gallery.
Queries are controlled feature-level perturbations of the corresponding public
exports rather than independently recollected maps. The strong perturbation
condition uses $0.30$ sparse-feature dropout and $0.10$ numeric noise.
Validation selects standardized Euclidean or cosine nearest-neighbor retrieval
separately for each representation. With 20 gallery identities, random Top-1
accuracy is $0.050$. We therefore interpret this experiment as
\emph{representation-level linkability}, not real-world cross-session home
re-identification.

\subsection{Representation-Aware Attacker Audit}

As defined in Section~\ref{Threat Model}, the adversary can use all information
exposed through the downstream representation. We evaluate two types of privacy
attacks: \emph{attribute-inference attacks}, which attempt to recover private
scene, object, or target attributes, and a \emph{representation-level
linkability attack}, which tests whether a perturbed representation can be
matched to the corresponding held-out scene.

We distinguish two evaluation cases. A \emph{restricted-feature attack} uses
only a subset or summary of the information contained in the exported
representation. A \emph{representation-aware attack} may additionally exploit
public fields and structural information available in that representation.
Representation awareness does not provide access to information inside the
protected local perception boundary; it permits fuller use of information
already available downstream.

This distinction matters because an attack does not necessarily exploit every
field or structural relationship contained in the representation it receives.
For example, an attribute-inference attack may operate only on aggregate
statistics even when the public export also contains spatial relationships,
graph structure, or additional geometric fields. Such an attack can therefore
underestimate residual leakage despite observing a representation that exposes
richer structure.

\noindent\textbf{Representation-aware feature mappings.}
The expanded mappings are constructed exclusively from public information.
For metric navigation, the expanded candidate includes the full public
free-space layout, absolute spatial-bin and point-distribution structure,
pairwise-distance information, and exported scalar counts. For normalized
navigation, expanded candidates use only normalized public structure, such as
normalized occupancy and point distributions, relative distances, and boundary
shape; removed global origin and metric scale are never reintroduced. For
topological navigation, public graph structure can be represented through
node/edge counts, degree and connectivity structure, component and cycle
statistics, clustering, endpoint relationships, and shortest-hop structure;
removed coordinates and metric edge lengths are not available.

For collision representations, expanded candidates incorporate public
inter-object arrangement and relational context in addition to the exported
exact or coarsened obstacle geometry. For object-goal representations, expanded
candidates combine the candidate-specific target reference with the complete
public planner layout and public target-relative spatial or structural
descriptors. No candidate receives raw RGB/RGB-D, hidden semantic targets,
scene identifiers, held-out labels, or other information retained exclusively
inside the protected local boundary.

The original restricted mapping is retained among the validation candidates.
Feature mapping and model family are selected using validation data only and
then frozen. Consequently, expanding the candidate attacker set does not imply
that the selected representation-aware model must obtain a higher held-out
score than the restricted model; choosing the larger test score would itself
constitute test-set selection.

We audit each evaluated representation--attack pair according to this
distinction. A \emph{representation--attack combination} denotes one privacy
attack applied to one particular exported representation. For example, room
inference from metric navigation and room inference from normalized navigation
are separate combinations because the adversary observes different public
representations.

Across the three robot tasks, the audit covers 33 representation--attack
combinations. Table~\ref{tab:attacker-audit} summarizes their distribution. For
7 combinations, the existing attack already consumed the relevant public
representation and required no feature expansion. The remaining 26 originally
used a restricted subset or summary, so broader representation-aware feature
variants were added to the validation-selection procedure.

\begin{table}[t]
\centering
\footnotesize
\caption{Representation-aware attacker audit. Each entry reports the total
number of representation--attack pairs, followed by already-complete and
expanded mappings in parentheses (complete/expanded).}
\label{tab:attacker-audit}
\setlength{\tabcolsep}{3pt}
\renewcommand{\arraystretch}{0.95}
\begin{tabularx}{\columnwidth}{
@{}p{0.22\columnwidth}
c
X
c@{}}
\toprule
\textbf{Task} &
\textbf{Reps.} &
\textbf{Privacy attacks} &
\textbf{Pairs (C/E)} \\
\midrule

Navigation &
$3$ &
Room; private-object; linkability &
$9\ (3/6)$ \\

Collision checking &
$3$ &
Room; private-object; object-category &
$9\ (3/6)$ \\

Object-goal &
$5$ &
Room; private-object; target-category &
$15\ (1/14)$ \\

\midrule
\textbf{Total} &
$\mathbf{11}$ &
&
$\mathbf{33\ (7/26)}$ \\
\bottomrule
\end{tabularx}
\end{table}

For navigation, the three linkability attacks already operated on the relevant
metric, normalized, and topological exports. For collision checking, the audit
additionally includes a semantic reference baseline, yielding three evaluated
collision representations; attacks on this baseline already used its relevant
semantic information. For object-goal execution, target-category inference from
the target-label representation likewise required no expansion because the
target category is directly exposed.

Table~\ref{tab:tfpd-primary-results} summarizes the primary held-out
privacy--utility results.

\begin{table*}[t]
\centering
\small
\caption{Primary held-out AI2-THOR privacy--utility results under
representation-aware attacks. Larger privacy-attack values indicate greater
measured residual leakage. All three navigation representations have mean path
ratio $0.898$ [95\% CI: $0.890, 0.907$].}
\label{tab:tfpd-primary-results}
\begin{tabular}{llcl}
\toprule
\textbf{Task} &
\textbf{Representation} &
\textbf{Utility} &
\textbf{Primary privacy attack [95\% CI]} \\
\midrule

Navigation &
$R_{\mathrm{metric}}$ &
success 1.000; ratio 0.898 &
linkability 0.970 [0.943, 0.992] \\

Navigation &
$R_{\mathrm{normalized}}$ &
success 1.000; ratio 0.898 &
linkability 0.532 [0.462, 0.612] \\

Navigation &
$R_{\mathrm{topological}}$ &
success 1.000; ratio 0.898 &
linkability 0.782 [0.738, 0.828] \\
\addlinespace

Collision &
anonymous exact geometry &
F1 1.000 [1.000, 1.000] &
object category 0.704 [0.652, 0.741] \\

Collision &
coarsened geometry &
F1 0.950 [0.937, 0.962] &
object category 0.556 [0.499, 0.592] \\
\addlinespace

Object goal &
target label &
success 0.995 [0.988, 1.000] &
target category 1.000 [1.000, 1.000] \\

Object goal &
target coordinate &
success 0.995 [0.988, 1.000] &
target category 0.094 [0.050, 0.125] \\

Object goal &
target region &
success 0.995 [0.988, 1.000] &
target category 0.077 [0.040, 0.102] \\

Object goal &
anonymous target ID &
success 0.995 [0.988, 1.000] &
target category 0.094 [0.051, 0.124] \\

Object goal &
target graph node &
success 0.995 [0.988, 1.000] &
target category 0.053 [0.022, 0.072] \\
\bottomrule
\end{tabular}
\end{table*}

\subsection{Navigation: Equivalent Utility and Different Spatial Leakage}

All three navigation representations achieve held-out success $1.000$
[95\% CI: $1.000,1.000$], path feasibility $1.000$, identical mean path
ratio $0.898$ [95\% CI: $0.890,0.907$], and zero planning failures across
1,000 held-out trials. They are therefore task-equivalent under the evaluated
navigation criteria while exposing substantially different privacy risks.

The clearest difference appears in representation-level linkability
(Table~\ref{tab:tfpd-primary-results}). Under strong perturbation, Top-1
linkability is $0.970$ for metric free space, $0.532$ for normalized free
space, and $0.782$ for topological free space, compared with the random
baseline of $0.050$. Coordinate normalization lowers linkability by $0.438$,
or 43.8 percentage points, relative to the metric representation without
changing the evaluated navigation utility.

Under the primary AI2-THOR representation-aware evaluation, the same
point-estimate ordering also appears for both navigation attribute-inference
risks:
\begingroup
\setlength{\abovedisplayskip}{3pt}
\setlength{\belowdisplayskip}{3pt}
\setlength{\abovedisplayshortskip}{3pt}
\setlength{\belowdisplayshortskip}{3pt}
\[
R_{\mathrm{normalized}}
<
R_{\mathrm{topological}}
<
R_{\mathrm{metric}}.
\]
\endgroup
For room inference, representation-aware macro-F1 is
$0.479$ [95\% CI: $0.275$, $0.643$] for normalized,
$0.558$ [$0.327$, $0.745$] for topological, and
$0.736$ [$0.542$, $0.896$] for metric free space, compared with the
most-frequent-class baseline of $0.100$. Thus, room context remains
substantially inferable after coordinate normalization and topological
abstraction. For private-object inference, macro-F1 is
$0.412$ [$0.333,0.474$] for normalized,
$0.567$ [$0.375,0.804$] for topological, and
$0.928$ [$0.722,1.000$] for metric free space, compared with the
most-frequent-class baseline of $0.444$. Metric free space therefore supports
strong private-object inference. The topological point estimate remains above
the baseline, whereas normalized free space does not exceed it.

These results show that privacy does not necessarily improve monotonically with
the amount of explicit spatial information removed. The topological
representation removes explicit coordinates, absolute metric scale, and metric
edge lengths, whereas the normalized representation retains relative metric
structure. Nevertheless, under the primary AI2-THOR evaluation, topological
free space has higher attack point estimates than normalized free space across
all three evaluated navigation privacy channels. Because the attribute-inference
confidence intervals overlap, we treat this as an observed point-estimate
ordering rather than a formal pairwise significance result.

The representation-aware re-evaluation further illustrates why the complete
public representation matters. For metric navigation, private-object macro-F1
increases from $0.405$ [95\% CI: $0.231,0.600$] under the restricted mapping
to $0.928$ [$0.722,1.000$] after validation selects the expanded public
structure. This increase reflects fuller use of the exported layout and public
scalar information rather than access to hidden state. Because the expanded
fields are introduced jointly, we do not attribute the increase to any single
field without a dedicated ablation.

Taken together, the primary navigation results show that removing explicit
semantics does not guarantee low residual leakage, task-equivalent exports need
not be privacy-equivalent, and stronger spatial abstraction does not induce a
universal monotonic reduction in measured leakage.

\paragraph{ProcTHOR navigation replication.}
We repeat the navigation experiment on a fixed set of 120 ProcTHOR-10K houses,
using 80 houses for training, 20 for validation, and 20 for held-out testing.
All 120 selected houses are usable, with no exclusions, substitutions, or
planning failures. ProcTHOR uses the AI2-THOR execution engine but provides a
separately generated distribution of complete, potentially multi-room houses.
We apply the same metric, normalized, and topological representation
constructions, navigation planner, perturbation parameters, and validation-only
retrieval selection without changing the representation definitions.

Across 1,000 held-out start--goal trials, all three representations again
achieve success $1.000$, path feasibility $1.000$, and identical mean path
ratio $0.879$ [95\% CI: $0.872,0.885$]. Despite this equivalent measured
utility, representation-level linkability remains strongly
representation-dependent: Top-1 is $1.000$ [95\% CI: $1.000,1.000$] for
metric free space, $0.573$ [$0.505,0.652$] for normalized free space, and
$0.497$ [$0.418,0.575$] for topological free space, compared with the random
baseline of $0.050$.

Thus, the central navigation finding replicates across the ProcTHOR house
distribution: task-equivalent exports remain privacy-inequivalent, and metric
free space remains substantially more linkable than normalized free space.
However, the relative ordering of normalized and topological linkability
reverses from the primary AI2-THOR evaluation, and their ProcTHOR confidence
intervals overlap. We therefore do not interpret either three-way ordering as
a universal property of representation abstraction. Instead, the replication
shows that the magnitude and precise ordering of residual leakage can depend on
the environmental distribution even when the broader privacy-inequivalence
phenomenon persists.

\subsection{Collision Checking: Geometry Leakage and Utility Tradeoff}

Collision representations are constructed from a common obstacle set before
crossing the representation-export boundary. Within the trusted local stage,
we exclude structural elements (\texttt{Floor}, \texttt{Wall},
\texttt{Ceiling}, \texttt{Window}, and \texttt{LightSwitch}) and retain
objects with valid axis-aligned bounding-box centers and sizes. Semantic labels,
object identities, and orientation are not exported. Anonymous exact geometry
represents each remaining obstacle as an axis-aligned cuboid
$b_i=(c_i,s_i)$, where $c_i$ denotes its center and $s_i$ its full size.
The coarsened representation applies
\[
Q(v)=\operatorname{round}(\operatorname{float}(v),1)
\]
independently to every available $x$, $y$, and $z$ component of both $c_i$
and $s_i$. Exact and coarsened representations therefore contain the same
ordered obstacle set and differ only in geometric precision.

Privacy and utility evaluations operate on these authoritative public
representations rather than reconstructing obstacle geometry from hidden scene
state. For collision checking, the exported cuboids are projected onto the
$x$--$z$ plane, public sizes are clamped to nonnegative values, rectangles
smaller than $0.03$ along either projected dimension are discarded, and
collision tests use a $0.05$ margin. No semantic labels, object identities, or
other hidden object information are available to the collision checker after
the export boundary.

Removing semantic labels and identities does not eliminate semantic inference
from geometry. Anonymous exact geometry yields object-category macro-F1
$0.704$ [95\% CI: $0.652,0.741$], compared with a most-frequent-class
baseline of only $0.00160$. Thus, substantial semantic category information
remains recoverable from geometry even though category labels are not
explicitly exported.

Relative to exact geometry, geometric coarsening lowers object-category inference to a macro-F1 of $0.556$ (95\% CI: $0.499$--$0.592$). This privacy reduction is not utility-free: collision-checking F1 decreases from $1.000$ (95\% CI: $1.000$--$1.000$) with exact geometry to $0.950$ (95\% CI: $0.937$--$0.962$) with coarsened geometry. Thus, coarsening weakens this semantic leakage channel but introduces a measurable loss in collision-checking utility.

The privacy benefit does not extend uniformly to the other measured channels.
Room-inference macro-F1 is $1.000$ [$1.000,1.000$] for both exact and
coarsened geometry, far above the room baseline of $0.100$. Private-object
inference is $0.844$ [$0.474,1.000$] for exact geometry and $0.928$
[$0.722,1.000$] for coarsened geometry, compared with the private-object
baseline of $0.444$.

We therefore do not interpret coarsening as a general privacy improvement. It
reduces object-category inference but incurs a measurable collision-utility
cost, leaves room inference unchanged at a high level, and increases the
private-object attack point estimate. This divergence motivates evaluating the
components of $\mathbf{L}_{\tau}(R)$ separately rather than assigning a single
privacy score to the representation.

\subsection{Object Goal: Target-Semantic Reduction and Residual Leakage}

TFPD first resolves the semantic target inside the protected local perception
boundary and then exports a task-functional reference to the already-resolved
target. We evaluate five representations: explicit target label, target
coordinate, target region, anonymous target ID, and planning-graph node. All
five achieve object-goal success $0.995$ [95\% CI: $0.988,1.000$].

The explicit target label makes target category directly recoverable and yields
macro-F1 $1.000$. Functional references substantially reduce this channel:
target-category macro-F1 is $0.094$ [$0.050,0.125$] for target coordinates,
$0.077$ [$0.040,0.102$] for the target region,
$0.094$ [$0.051,0.124$] for the anonymous ID, and
$0.053$ [$0.022,0.072$] for the graph node, compared with the
most-frequent-class target-category baseline of $0.00482$.

Thus, replacing the explicit semantic target with a task-functional reference
substantially reduces target-category disclosure while preserving the evaluated
task utility. The transformed references do not, however, eliminate residual
target-semantic inference under the evaluated attacks.

The representation-aware scene-level attacks reveal a different pattern.
Across all five target representations, room inference has the same held-out
point estimate of $0.644$, compared with a baseline of $0.100$. The
representation-specific 95\% CIs are
$[0.410,0.833]$ for target label,
$[0.414,0.825]$ for coordinates,
$[0.408,0.826]$ for region,
$[0.393,0.807]$ for anonymous ID, and
$[0.404,0.834]$ for graph node.

Private-object inference likewise has the same held-out point estimate of
$0.913$ across all five representations, compared with the baseline of
$0.444$. The corresponding CIs are
$[0.487,1.000]$ for target label,
$[0.608,1.000]$ for coordinates,
$[0.487,1.000]$ for region,
$[0.487,1.000]$ for anonymous ID, and
$[0.567,1.000]$ for graph node.

The target interface therefore strongly changes target-category leakage while
the two broader scene-level channels remain effectively unchanged under the
representation-aware public-layout attack. This provides direct empirical
motivation for the vector-valued residual-risk profile
$\mathbf{L}_{\tau}(R)$: a transformation can substantially improve one privacy
channel while leaving others far above their non-informative baselines.

One apparent anomaly illustrates the validation-only selection protocol.
Restricted target-label room inference was $0.796$, whereas the
validation-selected representation-aware mapping obtains $0.644$ on held-out
test. Representation awareness does not remove information; rather, the broader
candidate mapping is selected using validation data and is not guaranteed to
produce a monotonically larger test score. Selecting the larger of the two
held-out scores would introduce test-set selection.

These findings are specific to the evaluated object-goal criterion and do not
imply that the five target references are interchangeable for subsequent
manipulation or grasping tasks, which may require additional pose, orientation,
or grasp-relevant geometry.

\subsection{Controlled Synthetic Diagnostics}

We use a separately generated synthetic environment to isolate selected leakage
mechanisms under controlled representation changes. The benchmark contains
20 scenes across four room types, with a scene-disjoint split of 14 training and
6 test scenes. Object positions are randomized, while object dimensions are
fixed by type to create controlled category-dependent geometry. Unlike the
primary AI2-THOR protocol, the synthetic setting has no validation split.
Reported attacker results therefore represent diagnostic test-set upper
envelopes rather than validation-selected held-out estimates.

\paragraph{Geometry diagnostic.}
Removing explicit object labels does not eliminate semantic inference when
retained geometry remains predictive. Anonymous cuboids and the evaluated
geometric proxies achieve object-category macro-F1 $1.000$. Because dimensions
are fixed by type in this controlled generator, this result should not
be interpreted as real-world category-recovery performance; it isolates the
mechanism by which task-required geometry can encode removed semantics.

\paragraph{Target-reference diagnostic.}
An explicit target label yields target-category macro-F1 $1.000$, whereas the
evaluated task-functional references reduce it to $0.150$--$0.240$. These
representations still retain information needed to locate or reference the
target, so the diagnostic concerns target-semantic disclosure rather than
removal of all target information.

\paragraph{Spatial-linkability diagnostic.}
Under strong perturbation, Top-1 linkability is $0.428$ for the task-minimal
representation, $0.328$ for the coarse-grid representation, and $0.278$ for
the free-space-graph condition. Because the graph attacker uses only a reduced
aggregate summary rather than the complete graph, the $0.278$ result should not
be interpreted as evidence that topological representations are generally more
private.

Together, these diagnostics show that representation transformations can leave
leakage intact, reduce a particular leakage channel, or alter linkability
depending on what information remains. They provide mechanism-level support
for the primary AI2-THOR findings but are not used for direct numerical
comparison.
\section{Discussion}
\label{Discussion}

Our results suggest that robot privacy should be designed around task
requirements rather than only around predefined sensitive attributes. Different
robot tasks require different combinations of semantic, geometric, and spatial
information, and removing explicit fields does not necessarily remove the
information encoded by the remaining representation. TFPD therefore evaluates
candidate exports jointly by task utility, direct exposure, and multiple
residual inference risks rather than treating a representation as simply
``private'' or ``non-private.''

The experiments also show that privacy effects are channel-specific. A
transformation can reduce one leakage channel while leaving another unchanged
or increasing it, and stronger abstraction does not necessarily yield lower
leakage. This motivates task-specific, multi-risk representation design rather
than applying a single sanitization rule across downstream components.

TFPD currently profiles candidate representations rather than selecting among
them automatically at runtime. Future work can extend the framework to physical
robots, additional task requirements, and adaptive selection mechanisms that
account for utility, multiple privacy risks, and previously disclosed
information.

\paragraph{Limitations.}
Our primary evaluation uses AI2-THOR, and the navigation
task-equivalence/privacy-inequivalence finding also replicates on ProcTHOR.
However, ProcTHOR uses the same AI2-THOR execution engine, so this does not
constitute validation in an independent simulator or on a physical robot. The
changed normalized--topological linkability ordering across the two scene
distributions also shows that precise leakage rankings can depend on
environmental structure.

Representation-level linkability is measured using controlled perturbations of
the same public export rather than independent cross-session observations and
should not be interpreted as real-world home re-identification. Residual leakage
is empirical and attacker-dependent, and we do not evaluate repeated or jointly
observed exports across tasks or over time. Finally, because
representation-aware mappings introduce multiple public features jointly, we do
not attribute changes in leakage to individual features without dedicated
ablations.
\section{Conclusion}
\label{Conclusion}

This paper shows that domestic robot privacy cannot be evaluated only at the
level of raw sensing or explicit sensitive attributes. Representations exported
for downstream computation can themselves reveal private household information
through semantic, geometric, and spatial structure. We introduced
\emph{Task-Functional Perception Distillation} (TFPD), which places a
representation-export boundary between rich local perception and downstream
consumers, allowing task-functional representations to be released while richer
perceptual and semantic state remains local.

Our evaluation demonstrates that task functionality, representation abstraction,
and privacy leakage do not follow a simple monotonic relationship. In the
primary AI2-THOR evaluation, three task-equivalent navigation representations
achieve success $1.000$ and identical mean path ratio $0.898$, yet
representation-level linkability ranges from $0.532$ to $0.970$. The more
abstract topological representation also has higher attack point estimates than
coordinate normalization across all three evaluated navigation privacy channels.
A cross-distribution replication on ProcTHOR preserves the central finding that
task-equivalent representations can be privacy-inequivalent while changing the
relative ordering of normalized and topological linkability, showing that the
precise privacy ranking is not universal. For object-goal execution, replacing
an explicit target label with a target region reduces target-category macro-F1
from $1.000$ to $0.077$ while preserving success at $0.995$. For collision
checking, geometric coarsening reduces object-category macro-F1 from $0.704$
to $0.556$, but decreases collision F1 from $1.000$ to $0.950$ and does not
uniformly reduce other privacy risks.

These findings support a task-oriented, multi-risk view of robot privacy.
Attribute removal or stronger abstraction alone should not be treated as
evidence of privacy. Instead, privacy protection should account for the
information each robot task requires to cross the protected local boundary and
the private information that remains inferable from the complete exported
representation. TFPD therefore characterizes candidate exports jointly through
task utility, direct exposure, and multiple residual inference risks. More
broadly, our results show that task-equivalent representations can remain
privacy-inequivalent across different household-scene distributions, while
their precise leakage ordering can depend on the representation and
environment. This motivates task-specific, multi-risk privacy design rather
than assuming that any single representation transformation provides universal
privacy protection.



\clearpage
\bibliographystyle{IEEEtran}
\bibliography{reference}

\end{document}